\documentclass[11pt, a4paper, logo, twocolumn, copyright, nonumbering]{googledeepmind}
\pdfoutput=1

\usepackage[authoryear, sort&compress, round]{natbib}
\usepackage{adjustbox}
\usepackage{fancyvrb}
\usepackage{makecell}
\usepackage{multirow}
\usepackage{algorithm} 
\usepackage{algpseudocode}
\usepackage{amsmath} 
\usepackage[textwidth=18mm]{todonotes}
\usepackage{caption}
\usepackage{subcaption}
\usepackage{CJKutf8}
\usepackage{devanagari}
\usepackage{pgfplots}
\usepackage{tikz}

\pgfplotsset{compat=1.16}
\usetikzlibrary{shapes.misc, positioning}
\usetikzlibrary{arrows.meta}

\definecolor{blue200}{RGB}{174, 203, 250}
\definecolor{blue500}{RGB}{66, 133, 244}
\definecolor{blue700}{RGB}{25, 103, 210}
\definecolor{red200}{RGB}{246, 174, 169}
\definecolor{red500}{RGB}{234, 67, 53}
\definecolor{red700}{RGB}{197, 34, 31}
\definecolor{yellow200}{RGB}{253, 226, 147}
\definecolor{yellow500}{RGB}{251, 188, 4}
\definecolor{yellow700}{RGB}{242, 153, 0}
\definecolor{green200}{RGB}{168, 218, 181}
\definecolor{green500}{RGB}{52, 168, 83}
\definecolor{green900}{RGB}{13, 101, 45}

\newcommand{\greenbox}[1]{\begin{adjustbox}{bgcolor=green500}{\strut #1}\end{adjustbox}}

\newcommand{\enter}{\Pisymbol{psy}{191}}
\newcommand{\sentinel}[1]{\texttt{\color{Cerulean}<#1>}}
\newcommand{\fimstart}{\sentinel{|fim\_prefix|}}
\newcommand{\fimhole}{\sentinel{|fim\_middle|}}
\newcommand{\fimend}{\sentinel{|fim\_suffix|}}
\newcommand{\filesep}{\sentinel{|file\_separator|}}
\newcommand{\chinese}[1]{\begin{CJK*}{UTF8}{gbsn}#1\end{CJK*}}
\newcommand{\tchinese}[1]{\begin{CJK*}{UTF8}{bsmi}#1\end{CJK*}}
\newcommand{\korean}[1]{\begin{CJK*}{UTF8}{mj}#1\end{CJK*}}

\newcommand{\modelName}{CodeGemma}
\newcommand{\shortName}{CG}
\title{\modelName: Open Code Models Based on Gemma}

\renewcommand{\today}{2024-05-08}

\author[1]{\modelName{} Team, Google LLC}

\begin{abstract}
This paper introduces \modelName, a collection of specialized open code models built on top of Gemma, capable of a variety of code and natural language generation tasks.  We release three model variants. \modelName{} 7B pretrained (PT) and instruction-tuned (IT) variants have remarkably resilient natural language understanding, excel in mathematical reasoning, and match code capabilities of other open models. \modelName{} 2B is a state-of-the-art code completion model designed for fast code infilling and open-ended generation in latency-sensitive settings.
\end{abstract}

\begin{document}

\affil[1]{See \nameref{sec:contributions} section for full author list. Please send correspondence to \href{mailto:codegemma-team@google.com}{codegemma-team@google.com}.}

\maketitle

\section{Introduction}

We present \modelName, a collection of open code models based on Google DeepMind's Gemma models \citep{gemma2024}.

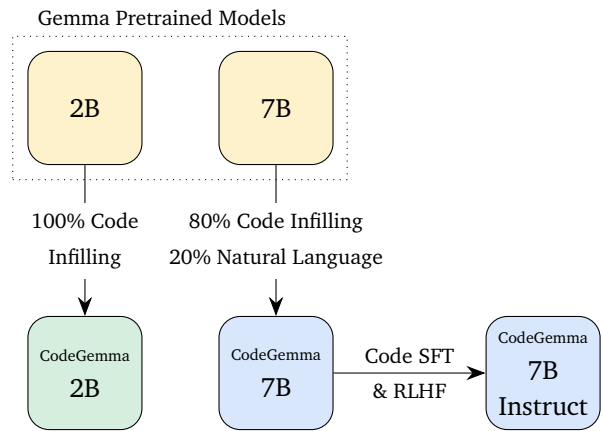
\begin{figure}[htb]
\centering
\begin{tikzpicture}
    \node (b) {\scriptsize Gemma Pretrained Models};
    \node (box) [draw, dotted, below=0cm of b.south west, anchor=north west, xshift=-0.2cm, minimum width=4.4cm, minimum height=1.9cm] {};
    \node (g2b) [draw, below=0.2cm of b.south west, anchor=north west, minimum size=1.5cm, fill=yellow500!20, rounded corners=8pt] {\small 2B};
    \node (g7b) [right=1cm of g2b.east, draw, minimum size=1.5cm, fill=yellow500!20, rounded corners=8pt] {\small 7B};
    \node (cg2b) [draw, below=2cm of g2b.south west, anchor=north west, minimum size=1.5cm, fill=green500!20, rounded corners=8pt, align=center] {\tiny\modelName\\\small 2B};
    \node (cg7b) [draw, below=2cm of g7b.south west, anchor=north west, minimum size=1.5cm, fill=blue500!20, rounded corners=8pt, align=center] {\tiny\modelName\\\small 7B};
    \node (cg7bit) [draw, right=2cm of cg7b.east, anchor=west, minimum size=1.5cm, fill=blue500!20, rounded corners=8pt, align=center] {\tiny\modelName\\\small 7B\\\small Instruct};
    \draw[arrows = {-Stealth[scale=1.5]}] (g2b) -- (cg2b) node [midway, fill=white, align=center] {\scriptsize 100\% Code\\\scriptsize Infilling};
    \draw[arrows = {-Stealth[scale=1.5]}] (g7b) -- (cg7b) node [midway, fill=white, align=center] {\scriptsize 80\% Code Infilling\\\scriptsize 20\% Natural Language};
    \draw[arrows = {-Stealth[scale=1.5]}] (cg7b) -- (cg7bit) node [midway, above, align=center] {\scriptsize Code SFT} node [midway, below, align=center] {\scriptsize \& RLHF};
\end{tikzpicture}
\caption{Both pretrained models are derived from corresponding Gemma pretrained models.}
\label{fig:lineage}
\end{figure}

Continuing from Gemma pretrained models, \modelName{} models are further trained on more than 500 to 1000 billion tokens of primarily code, using the same architectures as the Gemma model family. As a result, \modelName{} models achieve state-of-the-art code performance in both completion and generation tasks, while maintaining strong understanding and reasoning skills at scale. We release a 7B code pretrained model and a 7B instruction-tuned code model. Further, we release a specialized 2B model, trained specifically for code infilling and open-ended generation. The lineage of these models is depicted in Figure \ref{fig:lineage}.

We published the first collection (v1.0) on April 9\textsuperscript{th}, 2024 with all three models. A month later, on May 3\textsuperscript{rd}, 2024, we published a follow up (v1.1) to the pretrained 2B and instruction-tuned 7B models. Unless speed is critically important, we suggest v1.1 as it offers a well balanced quality improvement.

In this report, we provide an overview of the additions to Gemma, such as pretraining and instruction-tuning details for \modelName, followed by evaluations of all models across a wide variety of academic and real world tasks against similar models. Finally, we outline the areas in which \modelName{} excels and its limitations, followed by recommendations for using this model. Where applicable, we note the differences between v1.0 and v1.1.

\phantomsection
\section{Pretraining}

\phantomsection
\subsection{Training Data}
All \modelName{} v1.0 models are further trained on 500 billion tokens of primarily English language data from web documents, mathematics, and code. The 2B v1.1 model is trained on 1 trillion tokens. All 2B models are trained with 100\% code while the 7B models are trained with a 80\% code-20\% natural language mixture. Our code corpus comes from publicly available code repositories. Datasets are deduplicated and filtered to remove contamination of evaluation code and certain personal and sensitive data. In addition to the processing done for Gemma, we perform additional pretraining steps for code data.

\subsection{Preprocessing for Fill-in-the-Middle}

The pretrained \modelName{} models are trained using a method based on the fill-in-the-middle (FIM) task \citep{bavarian2022} with improvements that address the shortcomings cited in the original work as well as empirically-found systemic issues with existing FIM-trained models. The FIM rate is at 80\% in most models, except the pretrained 2B v1.1 where it is at 90\%. The relevant formatting control tokens are presented in Table \ref{tab:formatting_tokens}. The models are trained to work with both PSM (Prefix-Suffix-Middle) and SPM (Suffix-Prefix-Middle) modes. Figure \ref{fig:data_example} shows a sample snippet formatted in PSM. We make detailed FIM usage instructions in the \nameref{sec:inf_recommendations} section.

\begin{table}[htb]
    \setlength{\tabcolsep}{6pt}
    \centering
    \footnotesize
    \begin{tabular}{l c c}
    \toprule
    \textbf{Context} & \textbf{Relevant Token} \\
        \midrule
        \scriptsize{FIM prefix} & \fimstart \\
        \midrule
        \scriptsize{FIM middle} & \fimhole \\
        \midrule
        \scriptsize{FIM suffix} & \fimend \\
        \midrule
        \scriptsize{File separator} & \filesep \\
    \bottomrule
    \end{tabular}
    \caption{Formatting control tokens used for FIM task. Note that  \textbar{} is the standard pipe character (ASCII code 124).}
    \label{tab:formatting_tokens}
\end{table}

\begin{figure*}[htb]
    \footnotesize
    \rule{\linewidth}{1pt}
    \begin{Verbatim}[commandchars=\\\{\}]
path/to/the/first/file.py\greenbox{\enter}
\fimstart{}from typing import List\enter
\enter
def mean\_absolute\_deviation(numbers: List[float]) -> float:\enter
    """For a given list of input numbers, calculate Mean Absolute Deviation\enter
    around the mean of this dataset.\enter
    Mean Absolute Deviation is the average absolute difference between each\enter
    element and a centerpoint (mean in this case):\enter
    MAD = average | x - x\_mean |\enter
    >>> mean\_absolute\_deviation([1.0, 2.0, 3.0, 4.0])\enter
    1.0\enter
    """\enter
\fimend\fimhole    return sum(abs(x - mean) for x in numbers) / len(numbers)\enter
\filesep{}path/to/the/second/file.py\greenbox{\enter}
\fimstart...
\end{Verbatim}
    \rule{\linewidth}{1pt}
    \caption{Example code snippet in PSM mode. The green \greenbox{\enter} characters are part of the format, whereas uncolored \enter{} is from the source. The shown code sample is from HumanEval \citep{humaneval}.}
    \label{fig:data_example}
\end{figure*}

\subsection{Multi-file Packing}
Many downstream code-related tasks involve generating code based on a repository-level context as opposed to a single file. To improve model alignment with real-world applications, we create training examples by co-locating the most relevant source files within code repositories and best-effort grouping them into the same training examples. Specifically, we employ two heuristics: dependency graph-based packing and unit test-based lexical packing.

To construct the dependency graph, we first group files by repository. For each source file, we extract imports from the top N lines and perform suffix matching to determine the longest matching paths within the repository structure. We determine edge importance (a heuristic measure) between files, and remove unimportant edges to break cyclic dependencies (common in Python). We then calculate all-pairs shortest paths within the graph, where shorter distances signify stronger file relationships. Finally, we linearize the graph of files using a topological sort, selecting the next unparented node based on minimum distance to sorted nodes and using lexicographic order to break ties.

Files not covered by this dependency graph method are sorted alphabetically within their repository with unit tests packed next to their implementations (e.g. \texttt{TestFoo.java} beside \texttt{Foo.java}).

\section{Instruction Tuning}

Our training data consists of a combination of open-source math datasets and synthetically generated code, in addition to the finetuning datasets used by Gemma. By exposing the model to mathematical problems, we aim to enhance its logical reasoning and problem-solving skills, which are essential for code generation.

The instruction-tuned 7B v1.1 model differs from its 1.0 cousin in the reinforcement learning algorithm used (based on Gemma 1.1) and specifics of synthetic data generation. They all follow the general direction below.

\phantomsection
\subsection{Mathematics Datasets}
To enhance the mathematical reasoning capabilities of coding models, we employ supervised fine-tuning on a diverse set of mathematics datasets, including:

\begin{description}
\item[MATH Dataset] A collection of 12,500 challenging mathematical problems from competitions, providing step-by-step solutions for training models in answer derivation and explanation generation \citep{hendrycksmath2021}.

\item[GSM8k Dataset] A collection of 8,500 grade school math problems. This dataset tests the multi-step reasoning abilities of models, highlighting their limitations despite the simplicity of the problems \citep{cobbe_training_2021}.

\item[MathQA Dataset] A large-scale dataset of math word problems \citep{amini_mathqa_2019} with annotations built on top of the AQuA dataset \citep{ling_program_2017}.

\item[Synthetic Mathematical Data] A programmatically-generated dataset of algebraic problems used to improve ability to solve long algebra problems.
\end{description}

By leveraging these diverse datasets, we expose the model to a wide range of mathematical problems, increasing their ability to perform complex mathematical reasoning. Our training experiments indicate that these datasets significantly boost code generation performance.

\subsection{Coding Dataset}
Effectively instruction-tuning large language models for code generation tasks requires a substantial amount of question-answer pairs. We leverage synthetic code instruction data generation to create datasets used in the supervised-finetuning (SFT) and reinforcement learning from human feedback (RLHF) phase. We apply the following steps:

\begin{description}
\item[Example Generation] Following the approach outlined in the OSS-Instruct paper \citep{wei_magicoder_2023}, we generate a set of self-contained question-answer pairs.
\item[Post-Filtering] We filter question-answer pairs using an LLM tasked with evaluating the helpfulness and correctness of the generated question-answer pairs.
\end{description}

\section{Evaluation}
\label{sec:evals}

We evaluate \modelName{} for code completion and generation performance, as well as natural language understanding, with automated benchmarks across a variety of domains.

\phantomsection
\subsection{Infilling Capability}

\phantomsection
\subsubsection{HumanEval Infilling}

The \modelName{} models are trained for code completion purposes. We use the single-line and multi-line metrics in the HumanEval Infilling benchmarks introduced in \cite{incoder2023} to evaluate. Performance against other FIM-aware code models is shown in Table \ref{tab:fim_evals}.

\begin{table*}[b]
    \centering
    \begin{tabular}{l l c c c c}
    \toprule
        & & \multicolumn{2}{c}{Time (s)} & \multicolumn{2}{c}{Performance} \\
        \cmidrule(l{3pt}r{3pt}){3-4} \cmidrule(l{3pt}r{3pt}){5-6}
        & Model & Single & Multi & Single & Multi \\
    \midrule
        \multirow{4}{*}{\rotatebox[origin=c]{90}{2B class}} &
        \modelName{} & \textbf{543} & \textbf{8479} & 78.41\% & \textbf{51.44\%} \\
        & \modelName{} 1.1 & 771 & 9372 & 79.28\% & 50.99\% \\
        & DeepSeek Coder & 990 & 13138 & 79.96\%& 50.95\% \\
        & DeepSeek Coder Instruct & 5632 & 31505 & \textbf{81.41\%} & 37.35\% \\
        & StarCoder2 & 3665 & 20629 & 77.44\% & 47.65\% \\
    \midrule
        \multirow{6}{*}{\rotatebox[origin=c]{90}{7B class}} & \modelName{} & \textbf{1505} & 22896 & 76.09\% & 58.44\% \\
        & \modelName{} Instruct & 8330 & 49438 & 68.25\% & 20.05\% \\
        & \modelName{} Instruct 1.1 & 7579 & 46888 & 77.44\% & 23.66\% \\
        & Code Llama* & & & 74.10\% & 48.20\% \\
        & CodeQwen 1.5 & 7046 & 41345 & 32.72\% & 13.33\% \\
        & DeepSeek Coder & 1559 & \textbf{22387} & 85.87\% & \textbf{63.20\%} \\
        & DeepSeek Coder Instruct & 9500 & 53498 & \textbf{86.45\%} & 58.01\% \\
        & StarCoder2 & 8080 & 45459 & 81.03\% & 53.21\% \\
    \bottomrule
    \end{tabular}
    \caption{Single-line and multi-line code completion capability of \modelName{} compared to other FIM-aware code models \citep{codellama2024,qwen,guo2024deepseekcoder,lozhkov2024starcoder}. Time is the total number of seconds to obtain 128-token continuations per each HumanEval Infilling task (1033 tasks in single-line and 5815 multi-line). Measurements are done with HuggingFace's Transformers \citep{wolf2020huggingfaces} model implementations on \texttt{g2-standard-4} GCE instances with bfloat16 datatype and batch size of 1. 
    * Code Llama numbers are taken from \cite{codellama2024}.}
    \label{tab:fim_evals}
\end{table*}

We observe that our 2B pretrained model is an excellent well-rounded model for code completion use cases, where low latency is a critical factor. It performs on par with the other models while being, in many cases, nearly twice as fast during inference. We attribute this speedup to the base Gemma architectural decisions.

\subsubsection{Real-world Evaluation}

We validate our model's infilling abilities by masking out random snippets in code with cross-file dependencies, generating samples from the model, and retesting the code files with the generated snippets to show that it performs as expected, a similar approach to \cite{liu2023repobench} or \cite{ding2023crosscodeeval}. Due to our inclusion of very recently committed open source code, we do not use the evaluations directly, but use an internal version with the same testing methodology.

In addition to evaluating on offline evaluations, the model was tested within live coding environments to benchmark its performance against current Google completion models.

\subsection{Coding Capability}

\phantomsection
\subsubsection{Python Coding}

The canonical benchmarks used in coding evaluation are HumanEval  \citep{humaneval} and Mostly Basic Python Problems \citep{mbpp}. We present our results in Table \ref{tab:python-coding}.

\begin{table}[H]
    \centering
    \begin{tabular}{l c c c}
    \toprule
        Benchmark & HumanEval & MBPP \\
    \midrule
        2B-PT       & 31.1\%          & 43.6\% \\
        2B-PT 1.1   & \textbf{37.8\%} & \textbf{49.2\%} \\
        Gemma 2B PT & 22.0\%          & 29.2\% \\
    \midrule
        7B-PT       & 44.5\%          & \textbf{56.2\%} \\
        7B-IT       & 56.1\%          & 54.2\% \\
        7B-IT 1.1   & \textbf{60.4\%} & 55.2\% \\
        Gemma 7B PT & 32.3\%          & 44.4\% \\
    \bottomrule
    \end{tabular}
    \caption{Python coding capability of \modelName{} on de-facto coding benchmarks.}
    \label{tab:python-coding}
\end{table}

Compared to the base Gemma models \citep{gemma2024}, CodeGemma models perform significantly better on tasks from the coding domain.
 
\subsubsection{Multi-lingual Benchmarks}

BabelCode \citep{orlanski2023measuring} is used to measure the performance of \modelName{} on code generation across a variety of popular programming languages. Results are presented in Table \ref{tab:babelcode}.

\begin{table*}[htb]
    \centering
    \begin{tabular}{c l c c | c c c}
    \toprule
        & Language & 2B & \multicolumn{1}{c}{2B 1.1} & 7B & 7B-IT & 7B-IT 1.1 \\
    \midrule
        \multirow{7}{*}{\rotatebox[origin=c]{90}{HumanEval}}
        & C/C++         & \textbf{24.2\%} & 19.9\% & 32.9\% & 42.2\% & \textbf{46.6\%} \\
        & C\#           & 10.6\% & \textbf{26.1\%} & 22.4\% & 26.7\% & \textbf{54.7\%} \\
        & Go            & \textbf{20.5\%} & 18.0\% & 21.7\% & 28.6\% & \textbf{34.2\%} \\
        & Java          & 29.2\% & \textbf{29.8\%} & 41.0\% & 48.4\% & \textbf{50.3\%} \\
        & JavaScript    & 21.7\% & \textbf{28.0\%} & 39.8\% & 46.0\% & \textbf{48.4\%} \\
        & Kotlin        & 28.0\% & \textbf{32.3\%} & 39.8\% & \textbf{51.6\%} & 47.8\% \\
        & Python        & 21.7\% & \textbf{36.6\%} & 42.2\% & 48.4\% & \textbf{54.0\%} \\
        & Rust          & \textbf{26.7\%} & 24.2\% & 34.1\% & 36.0\% & \textbf{37.3\%} \\
    \midrule
        \multirow{7}{*}{\rotatebox[origin=c]{90}{MBPP}}
        & C/C++       & \textbf{47.1\%} & 38.9\% & 53.8\% & 56.7\% & \textbf{63.5\%} \\
        & C\#         & 28.7\% & \textbf{45.3\%} & 32.5\% & 41.2\% & \textbf{62.0\%} \\
        & Go          & \textbf{45.6\%} & 38.9\% & 43.3\% & 46.2\% & \textbf{53.2\%} \\
        & Java        & 41.8\% & \textbf{49.7\%} & 50.3\% & 57.3\% & \textbf{62.9\%} \\
        & JavaScript  & \textbf{45.3\%} & 45.0\% & 58.2\% & \textbf{61.4\%} & \textbf{61.4\%} \\
        & Kotlin      & 46.8\% & \textbf{49.7\%} & 54.7\% & 59.9\% & \textbf{62.6\%} \\
        & Python      & 38.6\% & \textbf{52.9\%} & 59.1\% & \textbf{62.0\%} & 60.2\% \\
        & Rust        & 45.3\% & \textbf{47.4\%} & 52.9\% & \textbf{53.5\%} & 52.3\% \\
    \bottomrule
    \end{tabular}
    \caption{Multi-lingual coding capability of \modelName{} (\shortName) on BabelCode-translated HumanEval and Mostly Basic Python Problems (MBPP) datasets. IT stands for instruction-tuned.}
    \label{tab:babelcode}
\end{table*}

\subsection{Language Capability}

\begin{figure*}[htb]
\centering
\begin{tikzpicture}
\begin{axis}[
scale only axis,
width=10cm,
height=7cm,
ybar,
ymin=0,
bar width=5pt,
xtick=data,
symbolic x coords={Gemma, PT, IT, IT11},
enlarge x limits=0.18,
xticklabels = {Gemma IT, \modelName{} PT, \modelName{} IT, \modelName{} IT 1.1},
x tick label style={font=\scriptsize},
y tick label style={font=\scriptsize},
legend cell align=left,
legend style={
    anchor=south,
    at={(0.5, 1.02)},
    legend columns=5,
    font=\tiny,
},
]
\addplot[fill=blue200] coordinates {
(Gemma, 83.2) (PT, 74.1) (IT, 82.1) (IT11, 83.1) 
};
\addplot[fill=red500] coordinates {
(Gemma, 81.2) (PT, 77.6) (IT, 76.3) (IT11, 75.6) 
};
\addplot[fill=yellow200] coordinates {
(Gemma, 81.2) (PT, 76.1) (IT, 72.0) (IT11, 72.5) 
};
\addplot[fill=blue700] coordinates {
(Gemma, 72.3) (PT, 64.2) (IT, 60.5) (IT11, 61.2) 
};
\addplot[fill=green200] coordinates {
(Gemma, 64.3) (PT, 60.0) (IT, 56.1) (IT11, 58.3) 
};
\addplot[fill=yellow700] coordinates {
(Gemma, 63.4) (PT, 51.5) (IT, 45.7) (IT11, 46.4) 
};
\addplot[fill=green900] coordinates {
(Gemma, 53.2) (PT, 75.6) (IT, 49.5) (IT11, 50.8) 
};
\addplot[fill=red200] coordinates {
(Gemma, 46.4) (PT, 44.2) (IT, 41.2) (IT11, 47.3) 
};
\addplot[fill=black!80] coordinates {
(Gemma, 24.3) (PT, 19.9) (IT, 20.9) (IT11, 22.3) 
};
\legend{Boolq, PIQA, TriviaQA, ARC-C, HellaSwag, MMLU, WinoGrande, GSM8K, MATH}
\end{axis}
\end{tikzpicture}
    \caption{Language capability comparison of \modelName{} and the instruction-tuned version of Gemma. Both Gemma and \modelName{} are in the 7B size class.}
    \label{fig:nlu}
\end{figure*}
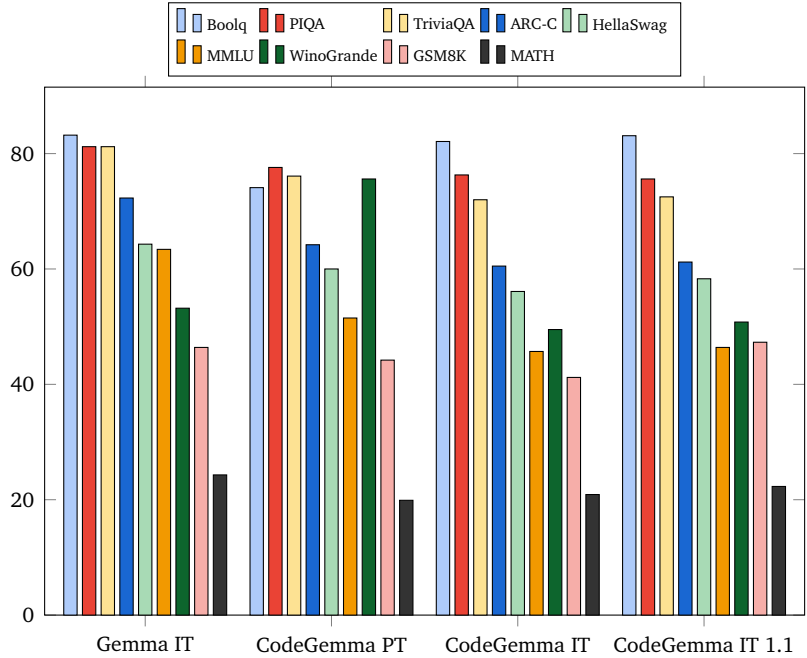

We evaluate performance on a variety of domains including question answering \citep{boolq, piqa, triviaqa, arc}, natural language \citep{zellers2019hellaswag, mmlu, winogrande} and mathematical reasoning \citep{gsm8k, hendrycksmath2021}. We present the results of our two 7B models next to the instruction-tuned Gemma 7B model in Figure \ref{fig:nlu}.

\modelName{} retains most of the same natural language capabilities seen in the base Gemma models. \modelName{} PT and IT both outperform Mistral 7B \citep{mistral} by 7.2\% and Llama-2 13B model \citep{llama2} by 19.1\% (numbers reported in \citealt{gemma2024}). Further, we compare scores for GSM8K and MATH in Table \ref{tab:math} from several code models in the 7B size class, and show that \modelName{} excels at mathematical reasoning compared to similarly sized models.

\begin{table}[H]
    \centering
    \begin{tabular}{l c c c}
    \toprule
        Model & GSM8K & MATH \\
    \midrule
        \modelName{} PT      & 44.2\% & 19.9\% \\
        \modelName{} IT      & 41.2\% & 20.9\% \\
        \modelName{} IT 1.1  & 47.3\% & 22.3\% \\
        Code Llama           & 13.0\% & \\
        DeepSeek Coder       & 43.2\% & 19.2\% \\
        StarCoder2           & 40.4\% & \\
    \bottomrule
    \end{tabular}
    \caption{Math reasoning capability of other code models in the same 7B size class. Results collected from \cite{codellama2024, guo2024deepseekcoder, lozhkov2024starcoder}.}
    \label{tab:math}
\end{table}

\section{Practical Considerations}

\modelName{} is tailored for practical use and deployment in latency-sensitive settings. The 2B model is considerably faster than all models in our comparison set, which is critical for latency-sensitive applications such as code completion. This speedup does not come with a significant, measured compromise in quality according to our evaluations — the 2B model performs as well or better compared to other open models in its class at code infilling tasks. Consequently, \modelName{} 2B is exceptionally suitable for utilization within Integrated Development Environments (IDEs), local environments, and other applications with memory constraints.

The 7B models, characterized by their strong performance, are general coding models that surpass the baseline Gemma models in terms of coding tasks while maintaining a high level of natural language comprehension. The larger memory requirement during inference renders these models particularly suitable for deployment in hosted environments and applications where model quality is of utmost importance.

The Responsible Deployment section in \cite{gemma2024} contains a thorough discussion about the limitations and benefits of using an open model.

\section{Inference Recommendations}
\label{sec:inf_recommendations}
For pretrained models, prompts should be formatted for code completion tasks such as function completion, docstring generation, and import suggestion. Figure \ref{fig:prompt_example} shows an example of a prompt format, where the file path is optional but recommended. The stopping strategy for model outputs should be chosen carefully to align with the deployment setting. The most straightforward method is to truncate upon generating a FIM sentinel token, as shown in Table \ref{tab:formatting_tokens}.

\begin{figure}[H]
    \footnotesize
    \rule{\linewidth}{1pt}
    \begin{Verbatim}[commandchars=\\\{\}]
path/file.py\greenbox{\enter}
\fimstart{}prefix\fimend{}suffix
\fimhole
\end{Verbatim}
    \rule{\linewidth}{1pt}
    \caption{Prompt in PSM mode. The carriage return \greenbox{\enter} is part of the format. There are no spaces after the suffix.}
    \label{fig:prompt_example}
\end{figure}

The same formatting as Gemma, with \sentinel{start\_of\_turn} and \sentinel{end\_of\_turn} tokens, can also prompt the instruction-tuned model.

\section{Conclusion}

We present a collection of open models specialized for coding applications, built on top of Gemma, an openly available family of language models \citep{gemma2024}. These models push the state of the art in code completion and generation, while retaining natural language capabilities from the base models.

The \modelName{} models presented in this report are highly capable language models designed for effective real-world deployment, optimized to be run in latency-constrained settings while delivering high-quality code completion on a variety of tasks and languages. We show that the lessons and technologies that built Gemini and Gemma are transferable to downstream applications, and we are excited to release these models to the broader community and to enable the applications which will be built on top of these models.

\clearpage

\section{Contributions and Acknowledgments}
\phantomsection
\label{sec:contributions}

\noindent\textbf{Core Contributors} \\
\chinese{赵赫日} (Heri Zhao) \\
\tchinese{許嘉倫} (Jeffrey Hui) \\
Joshua Howland \\
\selectlanguage{vietnamese}Nguyễn Thành Nam\selectlanguage{english}\footnote{Lead.} (Nam Nguyen) \\
\chinese{左斯琦} (Siqi Zuo)

\noindent\textbf{Contributors} \\
\tchinese{胡琪恩} (Andrea Hu) \\
Christopher A. Choquette-Choo \\
Jingyue Shen \\
Joe Kelley \\
{\dn E\322wEtj b\2sl} (Kshitij Bansal) \\
Luke Vilnis \\
Mateo Wirth \\
Paul Michel \\
Peter Choy \\
{\dn \3FEwEtk jofF} (Pratik Joshi) \\
Ravin Kumar \\
\includegraphics[trim=0 0.7em 0 0]{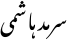} (Sarmad Hashmi) \\
{\dn \7{f}Bm ag\5vAl} (Shubham Agrawal)  \\
Zhitao Gong

\noindent\textbf{Product Management} \\
\selectlanguage{ukrainian}Євгенія Файн\selectlanguage{english} (Jane Fine) \\
Tris Warkentin

\noindent\textbf{Program Management} \\
Ale Jakse Hartman

\noindent\textbf{Executive Sponsors} \\
Bin Ni \\
Kathy Korevec \\
Kelly Schaefer \\
Scott Huffman

\noindent\textbf{Other Specialty Areas} \\
Special thanks and acknowledgments to these individuals for their assistance in respected areas:

\textbf{Central Support\ \ }
  David Huntsperger,
  Elisa Bandy,
  Emma Yousif,
  Glenn Cameron,
  James Freedman,
  Joe Fernandez,
  Josh Woodward,
  Keelin McDonell,
  Minh Giang

\textbf{Checkpoint Conversions\ \ }
  Austin Huang,
  Matthew Watson,
  Michael Butler,
  Michael Moynihan,
  Morgane Rivière,
  Phil Culliton,
  \chinese{单志昊} (Zhihao Shan)

\textbf{Ethics and Safety\ \ }
  Antonia Paterson,
  \includegraphics[trim=0 0.2em 0 0]{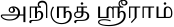} (Anirudh Sriram),
  Jenny Brennan,
  \chinese{帅杰} (Kevin Shuai),
  Ludovic Peran,
  \korean{김민} (Min Kim)

\textbf{Evaluations\ \ }
  {\dn aromA mh\?\306w\qa{d}{1}} (Aroma Mahendru),
  \includegraphics[trim=0 0.1em 0 0]{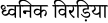} (Dhvanik Viradiya),
  {\dn gOrA\2g koEWyA} (Gaurang Kothiya),
  {\dn h\?\7{t}l pV\?l} (Hetul Patel),
  Jasmine George,
  Navneet Potti,
  {\dn s\2jnA \7{p}roEht} (Sanjana Purohit),
  {\dn u(kq\0 p\2\3B7wA} (Utkarsh Pandya)

\textbf{Gemma Model\ \ }
  Alek Andreev,
  Kathleen Kenealy,
  \includegraphics[trim=0 0.3em 0 0]{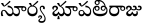} (Surya Bhupatiraju)

\textbf{Go-to-Market\ \ }
  Gabriel Rasskin,
  Kat Black,
  Lav Rai,
  Luiz Gustavo Martin,
  Manvinder Singh,
  Meg Risdal,
  \korean{박민우} (Minwoo Park),
  Nesh Devanathan,
  Nilay Chauhan,
  \includegraphics[trim=0 0.3em 0 0]{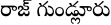} (Raj Gundluru)

\textbf{Partnering\ \ }
  Ankit Patel,
  Arthur Zucker,
  Bhavesh Pawar,
  Chenjie Luo,
  Christina Young,
  \chinese{孟东} (Dong Meng),
  Frederic Bastien,
  Izzy Putterman,
  Krzysztof Pawelec,
  \includegraphics[trim=0 0.2em 0 0]{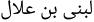} (Loubna Ben Allal),
  Marta Stepniewska-Dziubinska,
  Omar Sanseviero,
  Pedro Cuenca,
  Sharan Chetlur,
  {\dn v\4Bv \399wFvA-tv} (Vaibhav Srivastav),
  Yu-Hsuan Tseng

\textbf{Reinforcement Learning\ \ }
  Johan Ferret,
  Léonard Hussenot,
  Nino Vieillard,
  Olivier Bachem,
  Pier Giuseppe Sessa,
  Robert Dadashi,
  Sertan Girgin

\noindent\textbf{Team Acknowledgements} \\
Our work is made possible by the dedication and efforts of numerous teams at Google. We
would like to acknowledge the support from the following teams: AIDA, DevRel, Gemini Infrastructure, Gemini Safety, Gemma, Google Cloud, Google Research Responsible AI, Kaggle, Keras.

\bibliography{main}

\end{document}